\documentclass[11pt,a4paper, dvipsnames]{article}
\usepackage[hyperref]{arr}
\usepackage{amsmath}
\usepackage{xcolor}
\usepackage{times}
\usepackage{framed}
\usepackage{ragged2e}
\usepackage{bm}
\usepackage{soul}
\usepackage{latexsym}
\usepackage{subfigure}
\usepackage{amsthm}
\usepackage{graphicx}
\usepackage{float}
\usepackage{longtable}
\usepackage{booktabs} 
\usepackage{multirow}
\usepackage{amssymb}
\usepackage{dashbox}%
\usepackage{multirow}
\usepackage{textcomp}
\usepackage{tcolorbox}
\usepackage[ruled, vlined, linesnumbered]{algorithm2e}
\usepackage{mathtools}
\usepackage[ruled,vlined]{algorithm2e}
\usepackage{color, soul}
\usepackage{xcolor}
\definecolor{decentgrey}{RGB}{232,232,232}
\definecolor{mypattern}{RGB}{218,239,248}

\newtcbox{\pattern}{on line,colback=mypattern,colframe=white,size=fbox,arc=3pt, box align=base, before upper=\strut, top=-2pt, bottom=-2pt, boxrule=0pt}
\definecolor{mycolor1}{rgb}{0.341,0.627,0.839}
\definecolor{mycolor2}{rgb}{0.616,0.725,0.98}
\definecolor{mycolor3}{rgb}{0.678,0.851,0.969}
\definecolor{shadecolor}{rgb}{0.8,0.8,0.8}

\definecolor{mygreen}{RGB}{0,220,70}

\usepackage{microtype}
\usepackage[utf8]{inputenc}
\usepackage{cleveref}

\aclfinalcopy 

\crefname{section}{§}{§§}
\Crefname{section}{§}{§§}

\usepackage[T1]{fontenc}

\usepackage[utf8]{inputenc}

\title{Prompt-Learning for Fine-Grained Entity Typing}
\author{Ning Ding$^{1}$\thanks{\quad equal contribution}\hspace{0.5em}, Yulin Chen$^{3*}$, Xu Han$^{1,5}$, Guangwei Xu$^{2}$, \textbf{Pengjun Xie}$^{2}$, \\ \textbf{Hai-Tao Zheng}$^{3\dag}$, \textbf{Zhiyuan Liu}$^{1,5}$\thanks{\quad corresponding authors}, \textbf{ Juanzi Li$^{1}$}, \textbf{Hong-Gee Kim$^{4}$}\\
\textsuperscript{1}Department of Computer Science and Technology, Tsinghua University\\ 
\textsuperscript{2} Alibaba Group \  \textsuperscript{3} SIGS, Tsinghua University \  \textsuperscript{4} Seoul National University\\
\textsuperscript{5} State Key Lab on Intelligent Technology and Systems, Tsinghua University\\

\texttt{\{dingn18, yl-chen21, hanxu17\}@mails.tsinghua.edu.cn} \\

}
\date{}

\begin{document}
\maketitle

\begin{abstract}
As an effective approach to tune pre-trained language models (PLMs) for specific tasks, prompt-learning has recently attracted much attention from researchers. 
By using \textit{cloze}-style language prompts to stimulate the versatile knowledge of PLMs, prompt-learning can achieve promising results on a series of NLP tasks, such as natural language inference, sentiment classification, and knowledge probing. In this work, we investigate the application of prompt-learning on fine-grained entity typing in fully supervised, few-shot and zero-shot scenarios. 
We first develop a simple and effective prompt-learning pipeline by constructing entity-oriented verbalizer and templates and conducting masked language modeling. Further, to tackle the zero-shot regime, we propose a self-supervised strategy that carries out distribution-level optimization in prompt-learning to automatically summarize the information of entity types. Extensive experiments on three fine-grained entity typing benchmarks (with up to 86 classes) under fully supervised, few-shot and zero-shot settings show that prompt-learning methods significantly outperform fine-tuning baselines, especially when the training data is insufficient.


\end{abstract}

\section{Introduction}
\label{sec:intro}

In recent years, pre-trained language models (PLMs) have been widely explored and become a key instrument for natural language understanding~\cite{devlin2019bert,liu2019roberta} and generation~\cite{radfordimproving,raffel2020exploring}. By applying self-supervised learning on large-scale unlabeled corpora, PLMs can capture rich lexical~\cite{jawahar2019does}, syntactic~\cite{hewitt2019structural, wang2021cline}, and factual knowledge~\cite{petroni2019language} that well benefits downstream NLP tasks. Considering the versatile knowledge contained in PLMs, many efforts of researchers have been devoted to stimulating task-specific knowledge in PLMs and adapting such knowledge to downstream NLP tasks. 
Fine-tuning with extra classifiers has been one typical solution for adapting PLMs to specific tasks and achieves promising results on various NLP tasks~\cite{qiu2020pre,han2021pre}. 

\begin{figure}[!tbp]
    \centering
    \includegraphics[width = 1.0\linewidth]{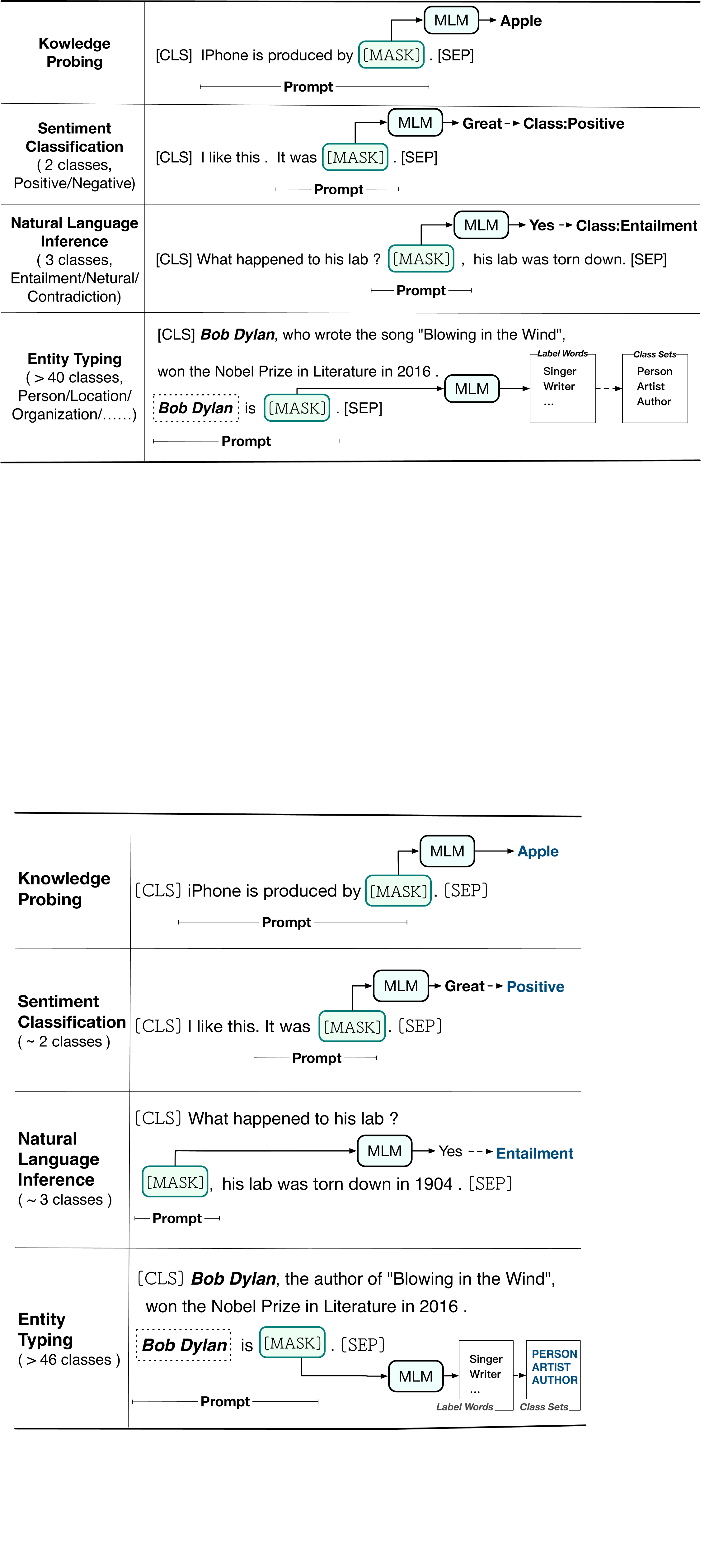}
    \caption{Examples of prompt-learning to stimulate the knowledge of PLMs by formalizing specific tasks as equivalent $cloze$-style tasks.}
    \label{fig:my_label}
\end{figure}


Some recent efforts on probing knowledge of PLMs show that, by writing some natural language prompts, we can induce PLMs to complete factual knowledge~\cite{petroni2019language}. GPT-3 further utilizes the information provided by prompts to conduct few-shot learning and achieves awesome results~\cite{brown2020language}. Inspired by this, prompt-learning has been introduced. As shown in Figure~\ref{fig:my_label}, in prompt-learning, downstream tasks are formalized as equivalent \textit{cloze}-style tasks, and PLMs are asked to handle these \textit{cloze}-style tasks instead of original downstream tasks. 
Compared with conventional fine-tuning methods, prompt-learning does not require extra neural layers and intuitively bridges the objective form gap between pre-training and fine-tuning. Sufficient empirical analysis shows that, 
either for manually picking hand-crafted prompts~\cite{liu2021gpt,han2021ptr} or automatically building auto-generated prompts~\cite{shin2020eliciting,gao2020making,lester2021power}, taking prompts for tuning models is surprisingly effective for the knowledge stimulation and model adaptation of PLMs, especially in the low-data regime.

Intuitively, prompt-learning is applicable to fine-grained entity typing, which aims at classifying marked entities from input sequences into specific types in a pre-defined label set.
We discuss this topic with a motivating example, ``\textit{He is from New York}''. 
By adding a prompt with a masking token $[\texttt{MASK}]$, 
the sentence becomes ``\textit{He is from New York. In this sentence, New York is $[\texttt{MASK}]$}''.
Due to the wealth of knowledge acquired during pre-training,
PLMs can compute a probability distribution over the vocabulary at the masked position, 
and a relatively higher probability with the word ``city'' than the word ``person''. In other words, with simple prompts, the abstract entity attributes contained in PLMs can be efficiently exploited, which is meaningful for downstream entity-related tasks.


In this work, we comprehensively explore the application of prompt-learning to fine-grained entity typing in fully supervised, few-shot and zero-shot settings. Particularly, we first introduce a naive pipeline, where we construct entity-oriented prompts and formalize fine-grained entity typing as a \textit{cloze}-style task. This simple pipeline yields promising results in our experiments, especially when supervision is insufficient. Then, to tackle the zero-shot scenario where no explicit supervision exists in training, we develop a self-supervised strategy under our prompt-learning pipeline. Our self-supervised strategy attempts to automatically summarize entity types by optimizing the similarity of the predicted probability distributions of paired examples in prompt-learning.

Three popular benchmarks are used for our experiments, including \textsc{Few-NERD}~\cite{ding2021few}, OntoNotes~\cite{MKJJ2R_2013}, BBN~\cite{Weischedel2005}. All these datasets have a complex type hierarchy consisting of rich entity types, requiring models to have good capabilities of entity attribute detection. Empirically, our method yields significant improvements on these benchmark datasets, especially under the zero-shot and few-shot settings. We also make an analysis and point out both the superiority and bottleneck of prompt-learning in fine-grained entity typing, which may advance further efforts to extract entity attributes using PLMs. Our source code and pre-trained models will be publicly available. 


\begin{figure*}[!htbp]
    \centering
    \includegraphics[width = 0.96\linewidth]{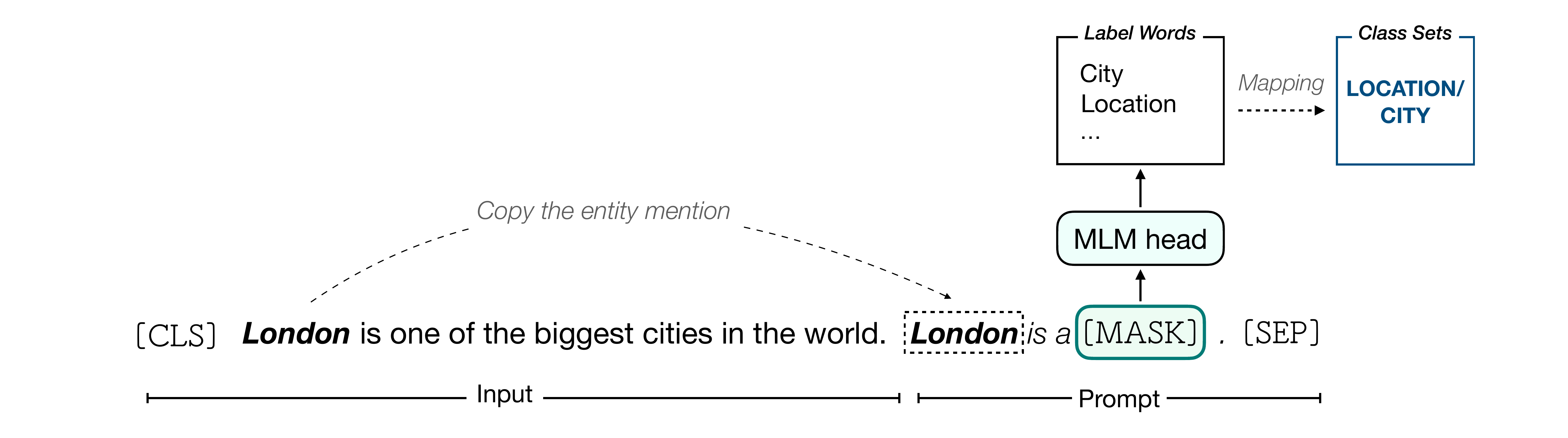}
    \caption{The illustration of prompt-learning for fine-grained entity typing with supervision. We take hard-encoding prompt strategy as an example in this figure.}
    \label{fig:plet}
\end{figure*}

\section{Background}
In this section, we first give a problem definition of the entity typing task (\cref{sec:def}), followed by an introduction of conventional vanilla fine-tuning (\cref{sec:vanilla}) and prompt-based tuning (\cref{sec:prompt}) with PLMs. 

\subsection{Problem Definition}
\label{sec:def}

The input of entity typing is a dataset $\mathcal{D} = \{x_1,...,x_n\}$ with $n$ sentences, and each sentence $x$ contains a marked entity mention $m$.
For each input sentence $x$, entity typing aims at predicting the entity type $y \in \mathcal{Y}$ of its marked mention $m$, where $\mathcal{Y}$ is a pre-defined set of entity types. Entity typing is typically regarded as a context-aware classification task. For example, in the sentence ``\textit{London} is the fifth album by the rock band Jesus Jones...'', the entity mention \textit{London} should be classified as \texttt{Music} rather than \texttt{Location}. In the era of PLMs, using pre-trained neural language models (e.g. BERT) as the encoder and performing model tuning for classifying types becomes a standard paradigm.



\subsection{Vanilla Fine-tuning}
\label{sec:vanilla}

In the vanilla fine-tuning paradigm of entity typing, for each token $t_i$ in an input sequence $x = \{\texttt{[CLS]}, t_1, \ldots, m, \ldots, t_T, \texttt{[SEP]}\}$ with a marked entity mention $m = \{t_i, \ldots, t_j\}$, the PLM  $\mathcal{M}$ produces its contextualized representation $\{\mathbf{h}_{\texttt{[CLS]}}, \mathbf{h}_1, \ldots, \mathbf{h}_T, \mathbf{h}_{\texttt{[SEP]}} \}$. Empirically, we choose the embedding of the $\texttt{[CLS]}$ token, $\mathbf{h}_{\texttt{[CLS]}}$, as the final representation that is fed into an output layer to predict the probability distribution over the label space 
\begin{equation}
    P(y \in \mathcal{Y}|s) =  \texttt{softmax}(\mathbf{W} \mathbf{h}_\texttt{[CLS]} +\mathbf{b}), 
\end{equation}
where $\mathbf{W}$ and $\mathbf{b}$ are learnable parameters. $\mathbf{W}$, $\mathbf{b}$ and all parameters of PLMs are tuned by maximizing the objective function $\frac{1}{n} \sum_{i=1}^n \log(P(y_i|s_i))$, where $y_i$ is the golden type label of $s_i$.




\subsection{Prompt-based Tuning}
\label{sec:prompt}
In prompt-based tuning, for each label $y \in \mathcal{Y}$,  we define a label word set $\mathcal{V}_y = \{w_1,\ldots,w_m\}$. $\mathcal{V}_y$ is a subset of the vocabulary $\mathcal{V}$ of the PLM $\mathcal{M}$, i.e., $\mathcal{V}_y \subseteq \mathcal{V}$. By taking the union of the dictionary corresponding to each label, we get an overall dictionary $\mathcal{V}^*$.
For example, in sentiment classification, we could map the label $y = \textsc{Positive}$ into a set $\mathcal{V}_y = \{\textit{great, good, wonderful...}\}$.
And another primary component of prompt-learning is a prompt template $T(\cdot)$, which modifies the original input $x$ into a prompt input $T(x)$ by adding a set of additional tokens at the end of $x$. Conventionally, a [\texttt{MASK}] token is added for PLMs to predict the missing label word $w \in \mathcal{V}^*$. Thus, in prompt-learning, a classification problem is transferred into a masked language modeling problem,
\begin{equation}
  p(y \in \mathcal{Y}|s)\!=\! p([\texttt{MASK}] \!=\! w \!\in\! \mathcal{V}_y | T(s)).
\end{equation}

\vspace{-0.1cm}
\section{Prompt-learning for Entity Typing: A Naive Pipeline}
\label{sec:plet}
\vspace{-0.1cm}

After transferred into masked language modeling, the prompt-learning method is applicable to learning and aggregating type information of entities.
In this section, we first introduce a naive but empirically strong baseline that utilizes prompts to extract entity types with explicit supervision, including the construction of label words (\cref{sec:label_words}), templates (\cref{sec:templates}) and training (\cref{sec:training}). And such a simple pipeline yields remarkable results on three benchmark datasets. Then we propose a self-supervised prompt-learning method that automatically learns type information from unlabeled data (\cref{sec:unplet}). 


\subsection{Label Words Set $\mathcal{V}^*$}
\label{sec:label_words}
\vspace{-0.1cm}

For fine-grained entity typing, datasets usually use hierarchical label space such as \textsc{Person/Artist} (\textsc{Few-NERD}) and \textsc{Organization/Party} (OntoNotes). In this case, we use all the words as the label words set $\mathcal{V}^*$ for this entity type. For example, $ y =  {\textsc{Location/City}} \rightarrow v = {\{\textit{location, city}\}}.$ And as the entity types are all well-defined nouns with clear boundaries, it is intuitive to expand the label words set $\mathcal{V}^*$ with obtainable related nouns. For example, in Related Words\footnote{\url{https://relatedwords.org}}, the top-10 related words of the label word $\textit{city}$ is ``\textit{metropolis}, \textit{town}, \textit{municipality}, \textit{urban}, \textit{suburb}, \textit{municipal}, \textit{megalopolis}, \textit{civilization}, \textit{downtown}, \textit{country}''. These words are strongly related to the class \textsc{City}, and they are hardly mapped to other entity types even under the same \textsc{Location} class, such as \textsc{Location/Mountain}, \textsc{Location/Island}, etc. 

In masked language modeling, we use confidence scores of all the words in $\mathcal{V}_y$ to construct the final score of the particular type $y$. That is, for an input $x$ (which is mapped to $T(x)$) and its entity type $y$ (which is mapped to $\mathcal{V}_y = \{w_1,...,w_m\}$), the conditional probability becomes
\begin{equation}
\label{eq:cond}
    P(y|x)\! =\! \frac{1}{m} \! \sum_j^m \lambda_j   P([\texttt{MASK}]\! =\! w_j|T(x)),
\end{equation}
where $\lambda_i$ is a parameter to indicate the importance of the current word $w_j \in \mathcal{V}_y$. 
Note that $\lambda_i$ could also be learnable or heuristically defined during the training procedure.

\subsection{Templates}
\label{sec:templates}

In this section, we construct entity-oriented prompts for the fine-grained entity typing task. We choose hard-encoding templates with natural language and soft-encoding templates with additional special tokens in our work. 

For the choice of hard-encoding templates, we do not use automatic searching methods for discrete prompts since the fine-grained entity typing task is clearly defined and the prompts are easily purposeful. We select simple declarative templates rather than hypernym templates to avoid grammartical errors.
In the template of hard encoding setting, we first copy the marked entity mention in $x$, then we add a few linking verbs and articles followed by the [\texttt{MASK}] token. With the marked entity mention $[\texttt{Ent}]$, we use the following templates: 
\begin{equation*}
\begin{split}
    &\text{T}_1 (x) = x.\ \pattern{[\texttt{Ent}]\text{ is }[\texttt{MASK}],} \\ 
    &\text{T}_2 (x) = x. \  \pattern{[\texttt{Ent}]\text{ is a }[\texttt{MASK}],} \\ 
    &\text{T}_3 (x) = x. \  \pattern{\text{In this sentence, }[\texttt{Ent}]\ \text{is a }[\texttt{MASK}],} \\ 
\end{split}
\end{equation*}
where \texttt{[Ent]} is the entity mention in $x$.
In \cref{sec:exp}, we report the the results of $\text{T}_3(\cdot)$.

We also adopt the soft-encoding strategy, which introduces some additional special tokens  $[\texttt{P}_1],...,[\texttt{P}_l]$ as the template, where $l$ is a pre-defined hyper-parameter.
The template begins with a delimiter [\texttt{P}] and a copy of the entity mention [\texttt{M}]. The complete template becomes:
\begin{equation*}
    \text{T}_4 (x)= x\  \pattern{[\texttt{P}]\  [\texttt{Ent}]\  [\texttt{P}$_1$],..., [\texttt{P}$_l$]\  [\texttt{MASK}],}
\end{equation*}
where each embedding  of prompts is randomly initialized and optimized during training. Intuitively, these special tokens can represent a cluster of words with similar semantics in the vocabulary.




\subsection{Training and Inference}
\label{sec:training}

The strategies of hard or soft encoding provide different initialization of templates, and they both can be parameterized by $\phi$ and optimized along with $\mathcal{M}$ during training. We train the pre-trained model $\mathcal{M}$ (parameterized by $\theta$) along with the additional prompt embeddings by using the cross-entropy loss function:
\begin{equation}
    \mathcal{L} = -  \sum \log P(y|x; \theta, \phi).
\end{equation}
For inference, we can directly use Eq.~\ref{eq:cond} to predict the label of the current input instance based on the predicted words of the \texttt{[MASK]} position.

This pipeline could be applied to entity typing task with explicit supervision, and it is effective even if the training data are insufficient, i.e., the few-shot scenario (\cref{exp_few}). Naturally, we consider a more extreme situation, that is, a scenario without any training data (zero-shot scenario). In this setting, if we directly use an additional classifier to predict the label, the result is equivalent to random guessing, because the parameters of the classifier are randomly initialized. If we use prompts to infer the label based on the predicted words, although its performance is significantly better than guessing, there will also be a catastrophic decline (\cref{sec:exp_zero}). At this time, a question emerges: \textit{``Is it possible for PLMs to predict entity types without any explicit supervision? ''} 
  

\begin{figure*}[h]
    \centering
    \includegraphics[width = 0.96\linewidth]{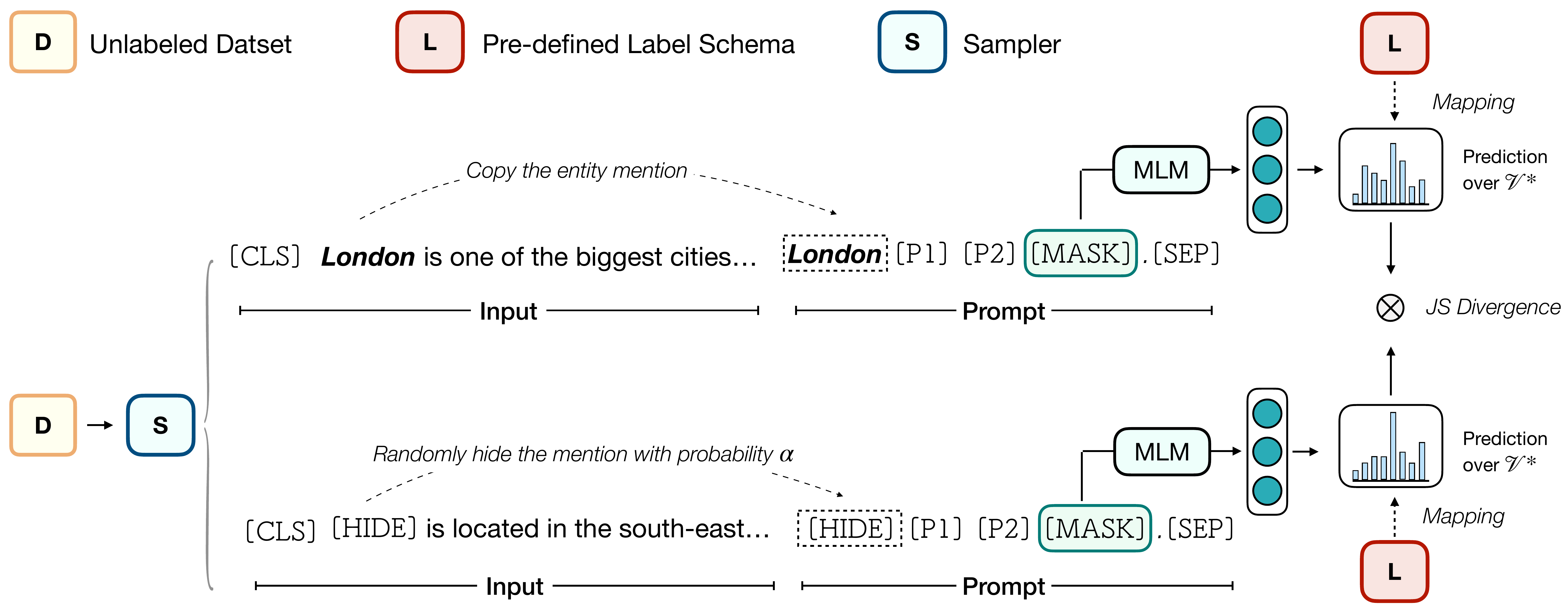}
    \caption{The illustration of self-supervised prompt-learning for fine-grained entity typing with unlabeled data and a pre-defined label set.   $\mathcal{V}^*$ denotes the label words projected from the input label set. Note that we only show the positive pair in this figure.}
    \label{fig:unplet}
\end{figure*}

\section{Self-supervised Prompt-learning for Zero-shot Entity Typing}
\label{sec:unplet}


With prompt-learning, the answer is yes, because in the pre-training stage, the contexts of entities have already implied the corresponding type information, which provides an advantageous initialization point for the prompt-learning paradigm. For example, in the input sentence with the $T_3(\cdot)$ template: ``Steve Jobs found Apple. \pattern{In this sentence, Steve Jobs is a \texttt{[MASK]}}''.
In our observations, the probability of PLMs predicting \textit{person} at the masked position will be significantly higher than the probability of \textit{location}.
And if we make reasonable use of this superior initialization point, it is possible for PLMs to automatically summarize the type information, and finally extract the correct entity type. 

\subsection{Overview}

In order to create conditions for PLMs to summarize entity types, we consider a self-supervised paradigm that optimizes the similarity of the probability distribution predicted by similar examples over a projected vocabulary $\mathcal{V}^*$.
To achieve that in prompt-learning, we need to (1) impose a limit on the prediction range of the model, so that only those words that we need, that is, words that express entity types, participate in the optimization of the gradient; (2) provide an unlabeled dataset, where entity mentions are marked without any types to allow the model to learn the process of inducing type information in a self-supervised manner.
 The inputs contain a pre-trained model $\mathcal{M}$, a pre-defined label schema $\mathcal{Y}$, and a dataset without labels $\mathcal{D} = \{x_1,...,x_n\}$ (entity mentions are marked without any types).
our goal is to make $\mathcal{M}$ capable to automatically carry out zero-shot entity typing after trained on $\mathcal{D}$ and $\mathcal{Y}$.
Using prompt-learning as the training strategy, we first construct a label words set $\mathcal{V}^*$ from $\mathcal{Y}$, and for each sentence $x$ in $\mathcal{D}$, we wrap it with hard-encoding template with a \texttt{[MASK]} symbol. The key idea is to make the prediction distributions of the same type of entities on $\mathcal{V}^*$ as similar as possible. In this way, we can perform contrastive learning  by sampling positive and negative examples, while ignoring the impact of other words that are not in $\mathcal{V}^*$ on optimization during the MLM process.

\subsection{Self-supervised Learning}

Although there are no labels in $\mathcal{D}$, we can still develop a sampling strategy based on a simple hypothesis, that is, \textit{same entities in different sentences have similar types}. For instance, we will sample two sentences contain ``Steve Jobs'' as a positive pair. Moreover, considering entity typing is context-aware, ``Steve Jobs'' could be \textit{entrepreneur}, \textit{designer}, \textit{philanthropist} in different contexts, we choose to optimize the similarity between distributions of the words over $\mathcal{V}^*$. This strategy not only softens the supervision, but also eliminates the impact of other words in self-supervised learning.

Particularly, we randomly sample $c$ positive pairs, i.e., sentence pairs that share one same entity mention, denoted as $\hat{\mathcal{D}_\text{pos}}$, and $c$ negative pairs, i.e., two sentences with different entity mentions marked, denoted as $\hat{\mathcal{D}_\text{neg}}$ from a large-scale entity-linked corpus $\mathcal{D}$. To avoid generating false negative samples, the negative samples are further restricted by a large dictionary that contains common entities and their type information. Only sentence pairs with entities of different types in the dictionary are selected as negative samples. Then we wrap them with hard-encoding $\text{T}_3(\cdot)$. To avoid overfitting of the entity names, we randomly hide the entity mention (in the original input and the template) with a special symbol [\texttt{Hide}] with a probability of $\alpha$. Empirically, $\alpha$ is set to 0.4. 

Since the impact of a pair of examples on training should be measured at the distribution level, we choose Jensen-Shannon divergence as a metric to assess the similarity of two distributions. 
Thus, in a sentence pair ($x$, $x'$), the similarity score of two representations of the the predictions $\mathbf{h}$ and $\mathbf{h}'$ of the [\texttt{MASK}] position is computed by:
\begin{equation}
s(\mathbf{h}, \mathbf{h}') = \texttt{JS}(P_{\mathcal{V}^*}(w|x), P_{\mathcal{V}^*}(w|x')),
\end{equation}
where \texttt{JS} is Jensen-Shannon divergence, $P_{\mathcal{V}^*}(w|x)$ and $P_{\mathcal{V}^*}(w|x')$ are probability distributions of the predicting token $w$ over $\mathcal{V}^*$ obtained by $\mathbf{h}$ and $\mathbf{h}'$.


As we attempt to make the predictions of the positive pairs similar, the objective is computed by:
\begin{equation}
\begin{split}
	\mathcal{L}= - \frac{1}{|\hat{\mathcal{D}_\text{pos}}|^2} \sum_{x \in \hat{\mathcal{D}_\text{pos}}} \sum_{x' \in \hat{\mathcal{D}_\text{pos}}} \log (1-s(\mathbf{h}, \mathbf{h'})) \\
	- \frac{1}{|\hat{\mathcal{D}_\text{neg}}|^2} \sum_{x \in \hat{\mathcal{D}_\text{neg}}} \sum_{x' \in \hat{\mathcal{D}_\text{neg}}} \gamma \log (s(\mathbf{h}, \mathbf{h'})),
\end{split}
\end{equation}
where $\gamma$ is a penalty term, because the assumption is loose in negative pairs. 
Overall, we use entity-linked English Wikipedia corpus as the raw data and generate about 1 million pairs of data each as $\hat{\mathcal{D}_\text{pos}}$ and $\hat{\mathcal{D}_\text{neg}}$.

\begin{table*}[!ht]
\centering
\scalebox{0.86}{
\begin{tabular}{llccccccccc}
\toprule
\multirow{2}{*}{\textbf{Dataset}} & \multirow{2}{*}{\textbf{\#Type}} &  \multicolumn{3}{c}{\textbf{Supervised }} & \multicolumn{3}{c}{\textbf{Few-shot}} & \multicolumn{3}{c}{\textbf{Zero-shot}} \\ \cmidrule(r){3-5} \cmidrule(r){6-8} \cmidrule(r){9-11}
                 &        &  $\left| \mathcal{D}_\text{train}  \right|$     & $\left| \mathcal{D}_\text{dev}  \right|$     &  $\left| \mathcal{D}_\text{test}  \right|$    & $\left| \mathcal{D}_\text{train}  \right|$     & $\left| \mathcal{D}_\text{dev}  \right|$     &  $\left| \mathcal{D}_\text{test}  \right|$   & $\left| \mathcal{D}_\text{train}  \right|$     & $\left| \mathcal{D}_\text{dev}  \right|$     &  $\left| \mathcal{D}_\text{test}  \right|$   \\  \cmidrule(r){1-1} \cmidrule(r){2-2} \cmidrule(r){3-5} \cmidrule(r){6-8} \cmidrule(r){9-11}
\multicolumn{1}{l|}{\textbf{Few-NERD}}                 & \multicolumn{1}{c|}{66} &     340,382      &    48,758     &    \multicolumn{1}{c|}{96,901}      &     66\textasciitilde1,056      &   $ = \left| \mathcal{D}_\text{train}  \right|$      &     \multicolumn{1}{c|}{96,901}    &      0     &    0    &     96,901    \\
\multicolumn{1}{l|}{\textbf{OntoNotes}}                       & \multicolumn{1}{c|}{86} &     253,239      &     2,200    &    \multicolumn{1}{c|}{8,962}      &     86\textasciitilde1,376      &  $=\left| \mathcal{D}_\text{train}  \right|$       &      \multicolumn{1}{c|}{8,962}   &       0    &     0    &     8,962    \\
\multicolumn{1}{l|}{\textbf{BBN}}                        & \multicolumn{1}{c|}{46} &     86,077      &     12,824    &     \multicolumn{1}{c|}{12,824}     &     46\textasciitilde736      & $=\left| \mathcal{D}_\text{train}  \right|$    &     \multicolumn{1}{c|}{12,824}    &     0      &    0     &    12,824
\\ \bottomrule
\end{tabular}}
\caption{Statistics of \textsc{Few-NERD}, OntoNotes and BBN from three experimental settings. It can be seen that for all three settings, the test sets are identical. For the training set of the few-shot setting, we report the summation from 1-shot to 16-shot.}
\label{table:stat}

\end{table*}

\begin{table*}[h]
\centering
\scalebox{0.86}{
\begin{tabular}{cccccccc}
\toprule
\multirow{2}{*}{\textbf{Shot}}                   & \multirow{2}{*}{\textbf{Metric}} & \multicolumn{2}{c}{\textbf{Few-NERD}}                & \multicolumn{2}{c}{\textbf{OntoNotes}}               & \multicolumn{2}{c}{\textbf{BBN}}                     \\ \cmidrule{3-4} \cmidrule{5-6} \cmidrule{7-8}
                                        &                         & Fine-tuning          & \textsc{Plet}        & Fine-tuning          & \textsc{Plet}        & Fine-tuning          & \textsc{Plet}        \\ \midrule
\multicolumn{1}{c|}{\multirow{3}{*}{1}}                      & \multicolumn{1}{c|}{Acc}                     &        8.94              &            43.87 {({\color{OliveGreen} +34.93})}         &           3.70           &        38.97 {({\color{OliveGreen} +35.27})}              &            0.80          &           40.70  {({\color{OliveGreen} +39.90})}          \\
\multicolumn{1}{c|}{\multirow{3}{*}{}}                      & \multicolumn{1}{c|}{MiF}   &         19.85             &         60.60  {({\color{OliveGreen} +45.75})}            &   18.98                   &          59.91  {({\color{OliveGreen} +40.93})} & 5.79         &          49.25   {({\color{OliveGreen} +43.46})}                               \\
\multicolumn{1}{c|}{\multirow{3}{*}{}}                      & \multicolumn{1}{c|}{MaF}    &        19.85  & 60.60  {({\color{OliveGreen} +40.75})}          &          19.43            &           61.42   {({\color{OliveGreen} +41.99})}         &             4.42         &         48.48   {({\color{OliveGreen} +43.06})}                          \\ 
\midrule
\multicolumn{1}{c|}{\multirow{3}{*}{2}}                      & \multicolumn{1}{c|}{Acc}             &      20.83                &            47.78   {({\color{OliveGreen} +26.95})}        &           7.27           &       39.19    {({\color{OliveGreen} +31.92})}            &            6.68          &           41.33   {({\color{OliveGreen} +34.65})}         \\
\multicolumn{1}{c|}{\multirow{3}{*}{}}                      & \multicolumn{1}{c|}{MiF}     &       32.67              &          62.09    {({\color{OliveGreen} +29.42})}         &   24.89                   &         61.09    {({\color{OliveGreen} +36.20})}          &         13.70             &          54.00    {({\color{OliveGreen} +40.30})}         \\
\multicolumn{1}{c|}{\multirow{3}{*}{}}                      & \multicolumn{1}{c|}{MaF}                &       32.67              &           62.09   {({\color{OliveGreen} +29.42})}         &         25.64             &         62.68   {({\color{OliveGreen} +37.04})}           &          13.23            &             51.97     {({\color{OliveGreen} +38.74})}     \\ 
\midrule
\multicolumn{1}{c|}{\multirow{3}{*}{4}}                      & \multicolumn{1}{c|}{Acc}          &      33.09                &           57.00   {({\color{OliveGreen} +23.91})}         &            11.15          &        38.39    {({\color{OliveGreen} +27.24})}           &          19.34            &           52.21   {({\color{OliveGreen} +32.87})}         \\
\multicolumn{1}{c|}{\multirow{3}{*}{}}                      & \multicolumn{1}{c|}{MiF}      &     44.14                 &           68.61     {({\color{OliveGreen} +24.47})}       &          27.69            &      59.81     {({\color{OliveGreen} +32.12})}            &         27.03             &              61.13  {({\color{OliveGreen} +34.10})}       \\
\multicolumn{1}{c|}{\multirow{3}{*}{}}                      & \multicolumn{1}{c|}{MaF}              &       44.14              &           68.61    {({\color{OliveGreen} +24.47})}        &             28.26        &        60.89   {({\color{OliveGreen} +32.63})}           &        24.69          &    58.91    {({\color{OliveGreen} +34.22})}               \\ 
\midrule
\multicolumn{1}{c|}{\multirow{3}{*}{8}}                      & \multicolumn{1}{c|}{Acc}             &      46.44                &           55.75    {({\color{OliveGreen} +9.31})}        &            18.37         &        39.37     {({\color{OliveGreen} +21.00})}          &          27.01            &           44.30      {({\color{OliveGreen} +17.29})}      \\
\multicolumn{1}{c|}{\multirow{3}{*}{}}                      & \multicolumn{1}{c|}{MiF}     &     57.76                 &           68.74    {({\color{OliveGreen} +10.98})}        &          38.16            &      57.97      {({\color{OliveGreen} +19.81})}           &         40.19           &              56.21   {({\color{OliveGreen} +16.02})}      \\
\multicolumn{1}{c|}{\multirow{3}{*}{}}                      & \multicolumn{1}{c|}{MaF}      &      57.76               &            68.74   {({\color{OliveGreen} +10.98})}        &            37.77         &       58.32    {({\color{OliveGreen} +20.55})}           &        39.50           &    55.15   {({\color{OliveGreen} +15.65})}                \\ 
\midrule
\multicolumn{1}{c|}{\multirow{3}{*}{16}}                      & \multicolumn{1}{c|}{Acc}             &      60.98               &           61.58    {({\color{OliveGreen} +0.60})}        &           32.26          &        42.29     {({\color{OliveGreen} +10.03})}          &          39.67            &          55.00  {({\color{OliveGreen} +15.33})}           \\
\multicolumn{1}{c|}{\multirow{3}{*}{}}                      & \multicolumn{1}{c|}{MiF}     &     71.59                 &           72.39   {({\color{OliveGreen} +0.80})}         &          51.40            &      60.79      {({\color{OliveGreen} +9.39})}           &         49.01             &              62.84   {({\color{OliveGreen} +13.83})}      \\
\multicolumn{1}{c|}{\multirow{3}{*}{}}                      & \multicolumn{1}{c|}{MaF}        &       71.59               &           72.39     {({\color{OliveGreen} +0.80})}       &             51.45         &         61.80    {({\color{OliveGreen} +10.35})}         &        47.09           &    62.38    {({\color{OliveGreen} +15.29})}               \\

\bottomrule
\end{tabular}}
\caption{Results of few-shot entity typing on \textsc{Few-NERD}, OntoNotes and BBN, all the methods use $\text{BERT}_\text{base}$ with same initialization weights as the backbone encoder. Training set and dev set have the same size. }
\label{tab:fewshot}
\end{table*}

\section{Experiments}
\label{sec:exp}

In this section, we conduct experiments to evaluate the effectiveness of our methods. We use FT to denote the BERT-based fine-tuning approach, \textsc{Plet} to denote the naive prompt-learning approach for entity typing in \cref{sec:plet}, and \textsc{Plet (S)} to denote the self-supervised prompt-learning approach in \cref{sec:unplet}.
Our experiments are carried out on fully supervised (\cref{sec:exp_sup}), few-shot (\cref{exp_few}) and zero-shot (\cref{sec:exp_zero}) settings on three fine-grained entity typing datasets. 

\subsection{Datasets}

We use three fine-grained entity typing datasets: \textsc{Few-NERD}, OntoNotes, and BBN.

 \textbf{\textsc{Few-NERD.}}\quad We use \textsc{Few-NERD}~\cite{ding2021few} as the main dataset, which has the following advantages: (1) \textsc{Few-NERD} is large-scale and fine-grained, which contains 8 coarse-grained and 66 fine-grained entity types. (2) \textsc{Few-NERD} is manually annotated, thereby we can precisely assess the capability of entity typing models. 
Specifically, we use the supervised setting of the dataset, \textsc{Few-NERD (SUP)}, and the official split of it to conduct our experiments. 

 \textbf{OntoNotes.}\quad We also use the OntoNotes 5.0 dataset~\cite{MKJJ2R_2013} in experiments. Following previous works for fine-grained entity typing, we adopt 86-classes version of OntoNotes, while each class has at most 3 levels of the type hierarchy. And the data split is identical to~\cite{shimaoka-etal-2017-neural}.  

 \textbf{BBN.}\quad BBN dataset is selected from Penn Treebank
corpus of Wall Street Journal texts and labeled by ~\cite{Weischedel2005}. We follow the version processed by~\cite{ren-etal-2016-afet}, and the data split by ~\cite{ren2016label}. The dataset contains 46 types and each type has a maximum type hierarchy level of 2.

\subsection{Experimental Settings}
The experiments are performed under three different settings to evaluate the effect of the prompt-learning method and semi-supervised training. In table \ref{table:stat}, we show the statistics of all the settings on the three datasets.

 \textbf{Supervised Setting.}\quad In a fully supervised setting, all training data are used in the training phase. FT and \textsc{Plet} are used to train the model. We run the experiments on all three datasets with BERT-base-cased backbone. Both hard and soft encodings are used for \textsc{Plet}. 

 \textbf{Few-shot Setting.}\quad In a few-shot setting, we randomly sample 1, 2, 4, 8, 16 instances for each entity type for training. We apply both FT and \textsc{Plet} methods with hard encoding on all the three datasets. 

 \textbf{Zero-shot Setting.}\quad In zero-shot setting, no training data with labels are available. The model is required to infer the entity type without any supervised training. Since fine-tuning is not applicable in this setting, we only conduct experiments on \textsc{Plet} and \textsc{Plet (S)}.


 \textbf{Metrics.}\quad In terms of evaluation metrics, we follow the widely used setting of ~\citet{Ling2012FineGrainedER}, which includes strict accuracy (Acc), loose macro F1-score (MaF) and loose micro F1-score (MiF) to evaluate the performances of models. The loose F1-score calculation concerns type labels by different granularities.

\subsection{Experimental Details}

We use $\text{BERT-base}$ ~\cite{devlin2019bert} as the backbone structures of our model and initialized with the corresponding pre-trained cased weights\footnote{\url{https://github.com/google-research/bert}}. 
The hidden sizes are 768, and the number of layers are 12. Models are implemented by Pytorch framework\footnote{\url{https://pytorch.org}}~\cite{paszke2019pytorch} and  Huggingface transformers\footnote{\url{https://github.com/huggingface/transformers}}~\cite{wolf2019huggingface}.
BERT models are optimized by AdamW~\cite{loshchilov2018decoupled} with the learning rate of 5e-5. The training batch size used is 16 for all models. In the supervised setting, each model is trained for 10 epochs and evaluated on the dev set every 2000 steps. In the few-shot setting, each model is trained for 30 epochs and evaluated every 10$\sim$50 steps, each time the evaluation is run for 200 steps. For the methods with hard-encoding, we report the experimental results of $T_3(\cdot)$. For the soft-encoding method, we report the results of $m=2$.
Experiments are conducted with CUDA on NVIDIA Tesla V100 GPUs. 

\subsection{Results of Fully Supervised Entity Typing}
\label{sec:exp_sup}
\begin{table}[h]
\centering
\scalebox{0.83}{
\begin{tabular}{lcccc}
\toprule
\multirow{2}{*}{\textbf{Dataset}}                      & \multirow{2}{*}{\textbf{Metric}} & \multicolumn{3}{c}{\textbf{Method}}                    \\ \cmidrule{3-5} 
                                              &                          & FT  & \textsc{Plet} ($\texttt {H}$) & \textsc{Plet} ($\texttt {S}$) \\ \midrule
\multicolumn{1}{l|}{\multirow{3}{*}{\textbf{Few-NERD}}} & \multicolumn{1}{c|}{Acc} & 79.75    &\textbf{79.90}         &        79.86       \\
\multicolumn{1}{c|}{}                         & \multicolumn{1}{c|}{MiF} & 85.74    &        \textbf{85.84}       &        {85.76}        \\
\multicolumn{1}{c|}{}                         & \multicolumn{1}{c|}{MaF} & 85.74    &        \textbf{85.84}        &        85.76   \\ \midrule
\multicolumn{1}{l|}{\multirow{3}{*}{\textbf{OntoNotes}}}    & \multicolumn{1}{c|}{Acc} & 59.71    &       {60.37}         &        \textbf{65.68}       \\
\multicolumn{1}{c|}{}                         & \multicolumn{1}{c|}{MiF} & 70.47    &        {70.78}      &      \textbf{74.53}       \\
\multicolumn{1}{c|}{}                         & \multicolumn{1}{c|}{MaF} & 76.57    &         76.42       &       \textbf{79.77}         \\ \midrule
\multicolumn{1}{l|}{\multirow{3}{*}{\textbf{BBN}}}     & \multicolumn{1}{c|}{Acc} & 62.39    &         \textbf{65.92}        &       63.11          \\
\multicolumn{1}{c|}{}                         & \multicolumn{1}{c|}{MiF} & 68.88    &       \textbf{71.55}          &        68.68        \\
\multicolumn{1}{c|}{}                         & \multicolumn{1}{c|}{MaF} & 67.37    &       \textbf{70.82}         &        67.81      \\ \bottomrule
\end{tabular}}
\caption{Fully supervised entity typing results. FT denotes the vanilla fine-tuning method, (\textsc{H}) denotes the hard-encoding strategy and  (\textsc{S}) denotes the soft-encoding strategy. All the methods use $\text{BERT}_\text{base}$ with same initialization weights as the backbone encoder.}
\label{tab:full-sup}
\end{table}

The results on all three datasets across different models are reported in Table~\ref{tab:full-sup}. Overall, the prompt-based methods have shown certain improvements comparing to directly fine-tuned models. It shows that the prompt-based method does help with capturing entity-type information from a given context. 

It is also observed that the magnitude of the improvement and the preference of prompt encoding strategy may vary with different datasets. The prompt-based method seems less effective on \textsc{Few-NERD} dataset than the other two. It indicates that the effect of the prompt-based method partially depends on the characteristics of the dataset and that different prompt designs may suit different data. Specifically, \textsc{Few-NERD} is manually annotated and contains much less noise than the other two datasets, benefiting the FT method to learn classification with an extra linear layer.
Moreover, for the OntoNotes dataset, soft encoding significantly outperforms hard encoding, while for the other two datasets the effect seems reversed. 

\subsection{Results of Few-shot Entity Typing}
\label{exp_few}

Table~\ref{tab:fewshot} shows the results on few-shot entity typing. It is shown that prompt-based model outperforms fine-tuning by a large margin under few-shot setting, especially when only 1 $\sim$ 2 training instances per type are available. It should be noted that for OntoNotes and BBN datasets, sampling 16 instances for each entity type already amounts to over 0.5\% of the total training data. Meanwhile, some of the data in BBN are distantly-supervised and are potentially erroneous. It brings more randomness to few-shot training. The results support the idea that a well-designed prompt has much potential in mining the learned knowledge in pre-trained models and thus yields better performance in few-shot settings. The results also indicate that even when the number of entity types is large (46 $\sim$ 86), the superiority of prompt-learning still holds.


\subsection{Results of Zero-shot Entity Typing}
\label{sec:exp_zero}


\begin{table}[]
\centering
\scalebox{0.88}{
\begin{tabular}{lccc}
\toprule
\multirow{2}{*}{\textbf{Dataset}}                      & \multirow{2}{*}{\textbf{Metric}} & \multicolumn{2}{c}{\textbf{Method}}                    \\ \cmidrule{3-4}
 &  & \multicolumn{1}{c}{\textsc{Plet}} & \multicolumn{1}{c}{\textsc{Plet (S)}} \\ \midrule 
\multicolumn{1}{l|}{\multirow{3}{*}{\textbf{Few-NERD}}} & \multicolumn{1}{c|}{Acc} & 17.55 & 23.99 {({\color{OliveGreen} +6.44})}\\
\multicolumn{1}{l|}{} & \multicolumn{1}{c|}{MiF} & 28.39 & 47.98 {({\color{OliveGreen} +19.59})}\\
\multicolumn{1}{l|}{} & \multicolumn{1}{c|}{MaF} & 28.39 & 47.98 {({\color{OliveGreen} +19.59})}\\ 
\midrule
\multicolumn{1}{l|}{\multirow{3}{*}{\textbf{OntoNotes}$^{\ddag}$}} & \multicolumn{1}{c|}{Acc} & 25.10 & 28.27 {({\color{OliveGreen} +3.17})}\\
\multicolumn{1}{l|}{} & \multicolumn{1}{c|}{MiF} & 33.61 & 49.79 {({\color{OliveGreen} +16.18})}\\
\multicolumn{1}{l|}{} & \multicolumn{1}{c|}{MaF} & 37.91 & 49.95 {({\color{OliveGreen} +12.04})}\\ 
\midrule
\multicolumn{1}{l|}{\multirow{3}{*}{\textbf{BBN}}} & \multicolumn{1}{c|}{Acc} & 55.82 & 57.79 {({\color{OliveGreen} +1.97})}\\
\multicolumn{1}{l|}{} & \multicolumn{1}{c|}{MiF} & 60.64 & 63.24 {({\color{OliveGreen} +2.60})}\\
\multicolumn{1}{l|}{} & \multicolumn{1}{c|}{MaF} & 59.99 & 64.00 {({\color{OliveGreen} +4.01})}\\ 
\bottomrule
\end{tabular}}
\caption{Results of zero-shot entity typing on \textsc{Few-NERD}, OntoNotes, and BBN. $^{\ddag}$ means that we remove the ``\texttt{Other}'' class during testing.
\textsc{Plet} denotes the prompt-learning pipeline and \textsc{Plet (S)} denotes self-supervised prompt-learning, both methods use the $\text{BERT}_\text{base}$ as the backbone encoder. }
\label{tab:zeroshot}
\end{table}


\begin{figure*}[htbp]

\centering

\subfigure[Zero-shot prediction distribution on \textsc{Org-SportsLeague}.]{
\begin{minipage}[t]{0.5\linewidth}
\centering
\includegraphics[width=0.98\linewidth]{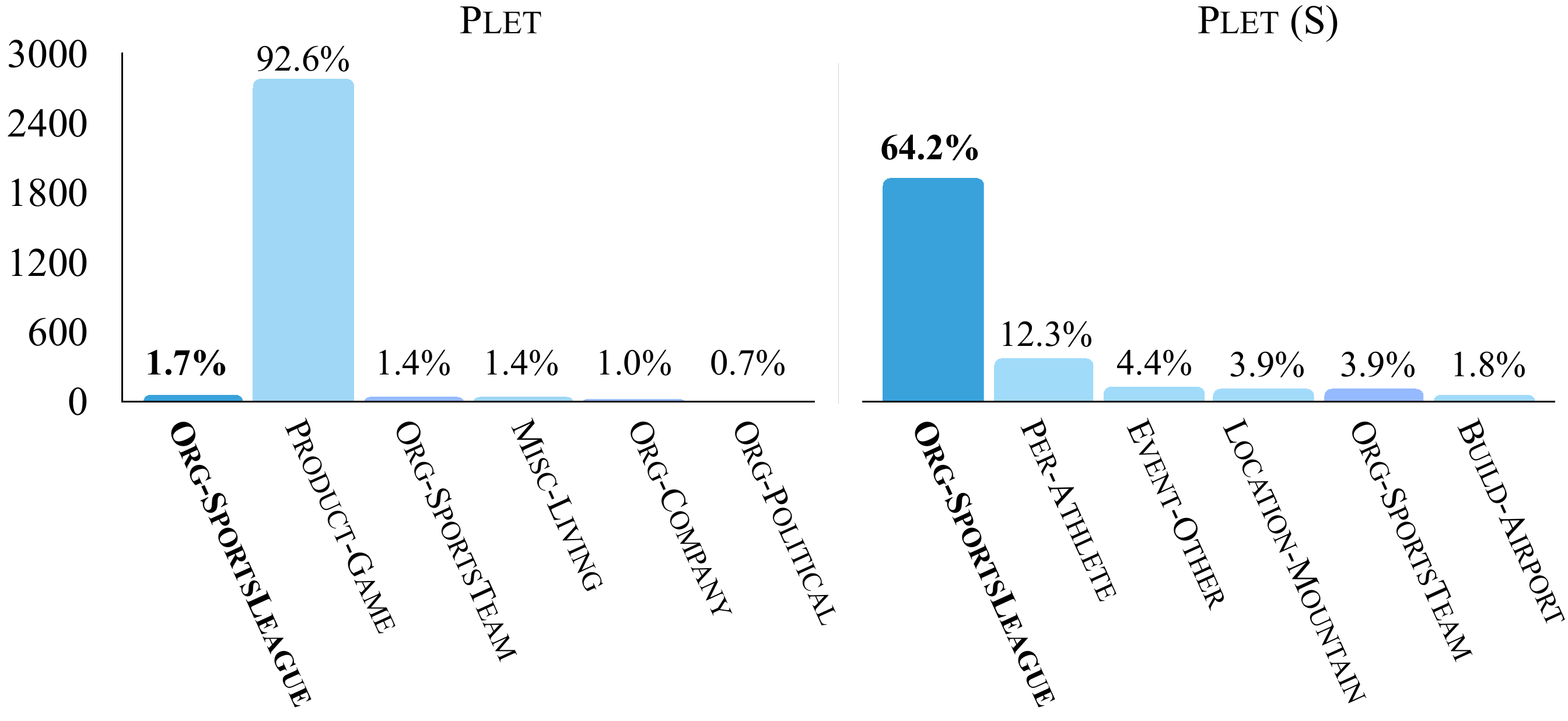}
\end{minipage}%
}%
\subfigure[Zero-shot prediction distribution on \textsc{Event-Attack}.]{
\begin{minipage}[t]{0.5\linewidth}
\centering
\includegraphics[width=0.98\linewidth]{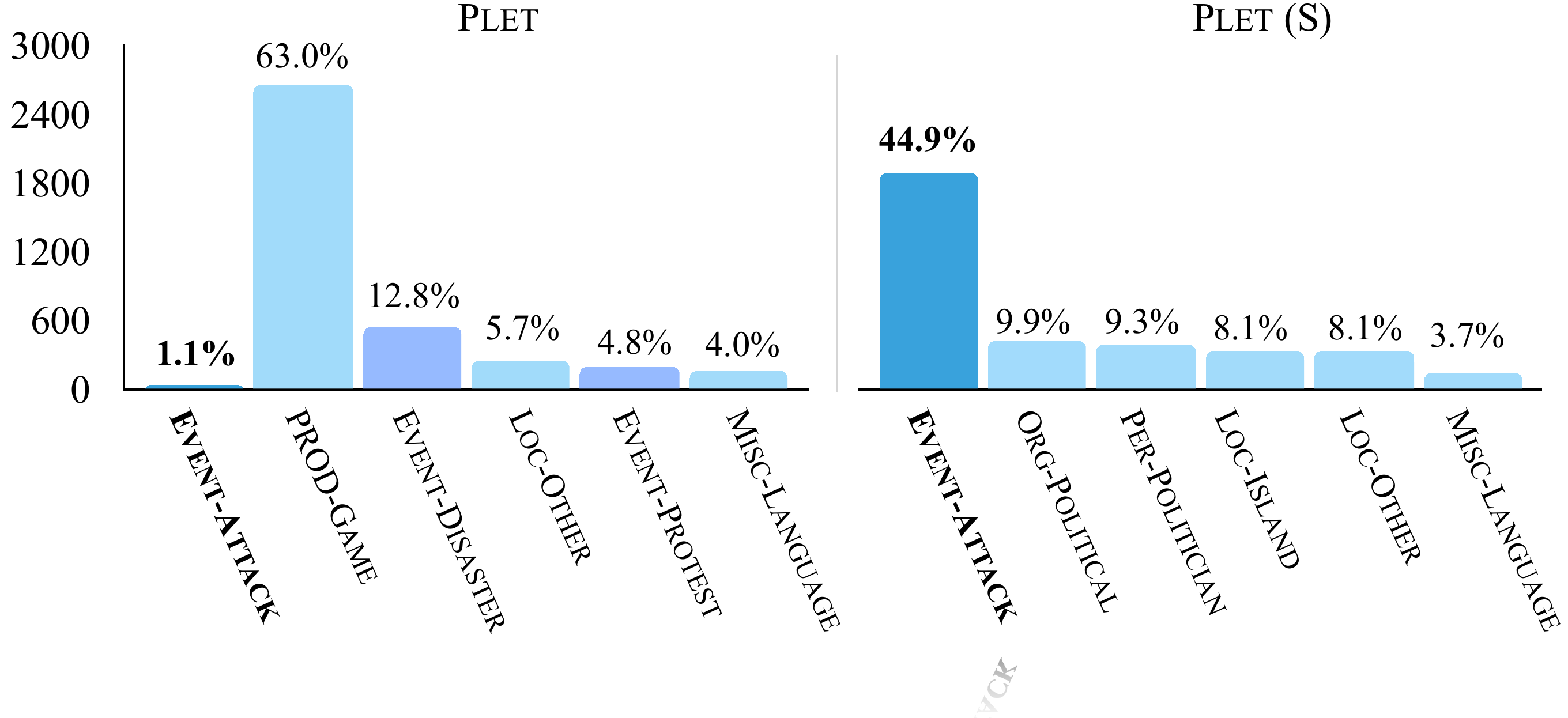}
\end{minipage}%
}%

\subfigure[Zero-shot prediction distribution on  \textsc{Misc-Currency}.]{
\begin{minipage}[t]{0.5\linewidth}
\centering
\includegraphics[width=0.98\linewidth]{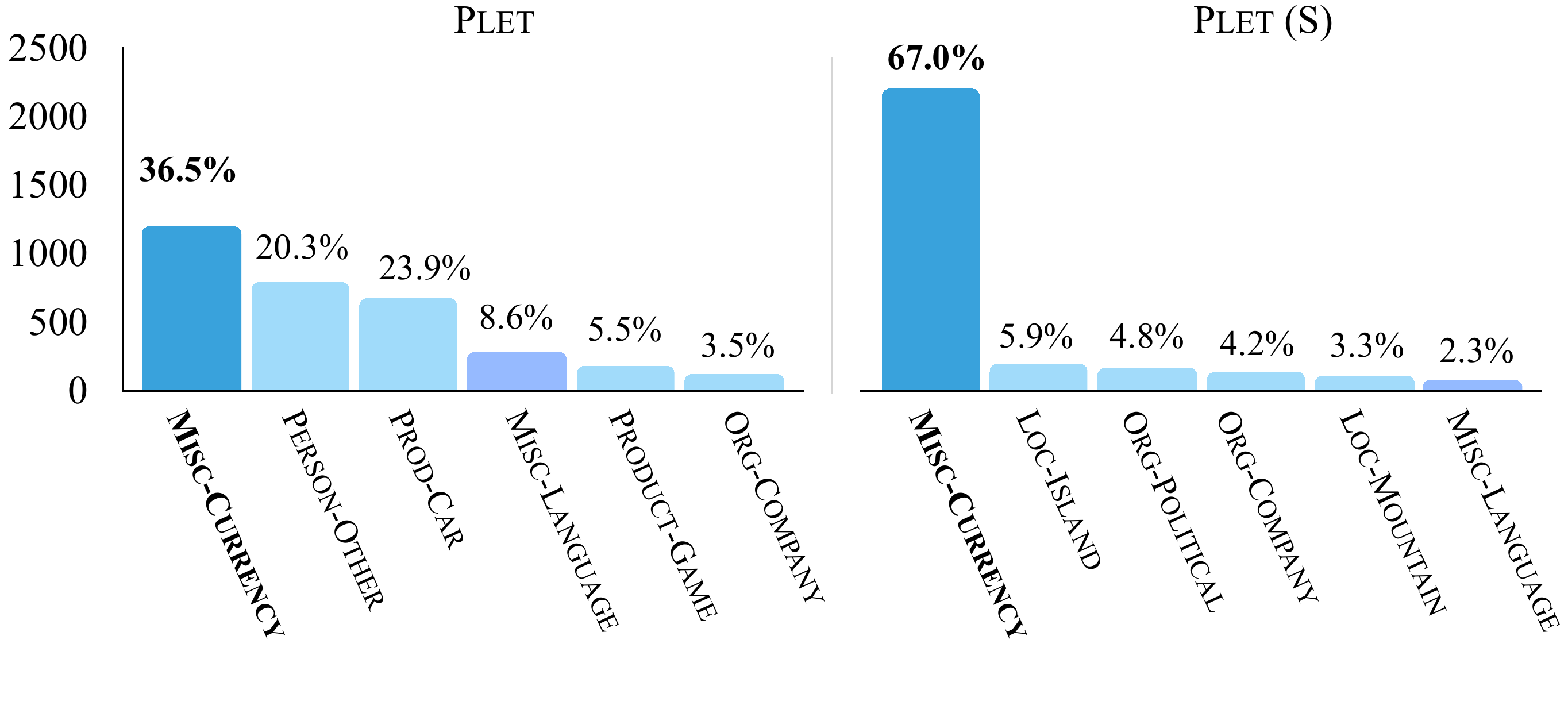}
\end{minipage}
}%
\subfigure[Zero-shot prediction distribution on \textsc{Loc-Mountain}.]{
\begin{minipage}[t]{0.5\linewidth}
\centering
\includegraphics[width=0.98\linewidth]{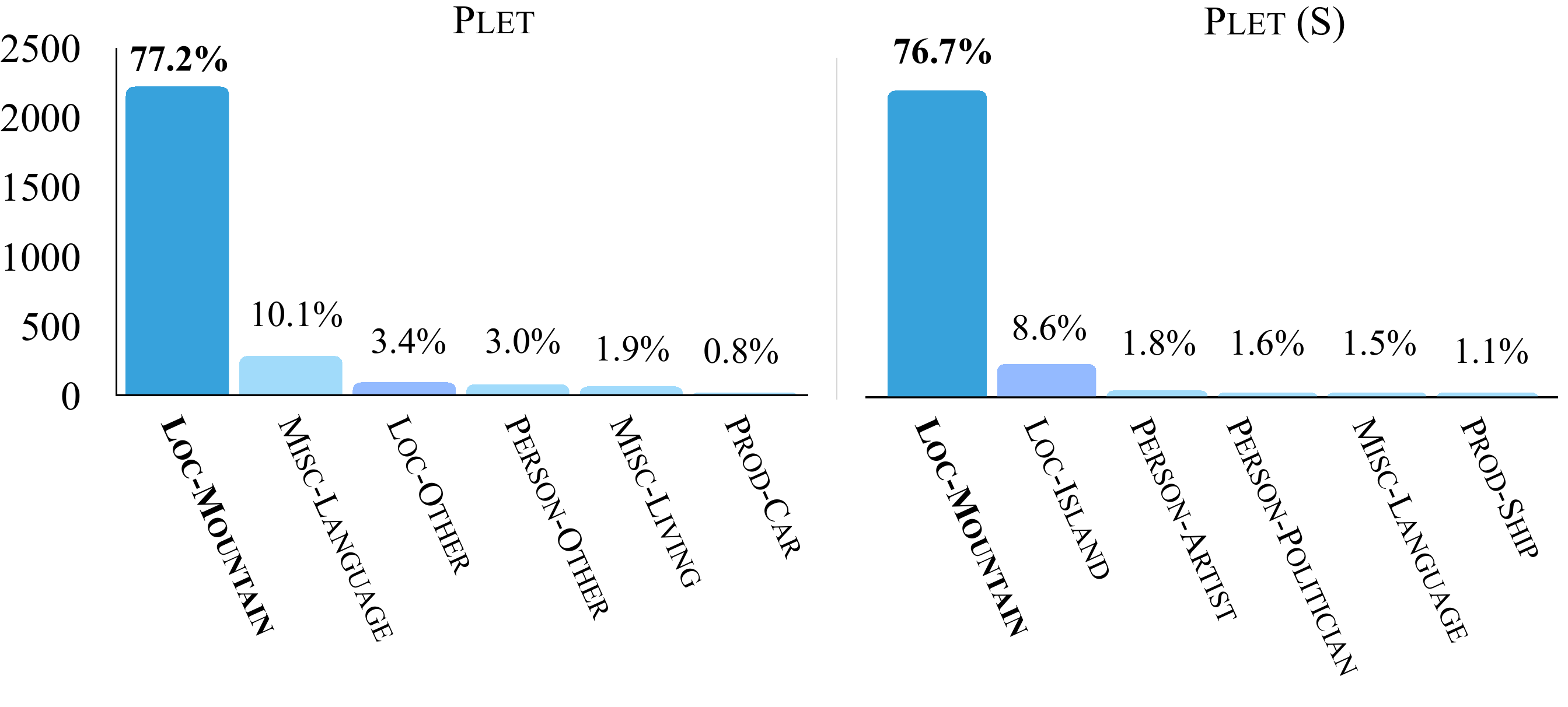}
\end{minipage}
}%

\centering

\caption{ Zero-shot prediction distribution on four types in \textsc{Few-NERD}, in each subgraph, the left part illustrates the results of \textsc{Plet} and the right part shows the results of \textsc{Plet (S)}. 
\fcolorbox{black}{mycolor1}{\rule{0pt}{4pt}\rule{4pt}{0pt}} denotes the correct predictions,  \fcolorbox{black}{mycolor2}{\rule{0pt}{4pt}\rule{4pt}{0pt}} denotes the wrong predictions with correct coarse-grained types, and \fcolorbox{black}{mycolor3}{\rule{0pt}{4pt}\rule{4pt}{0pt}} denotes the wrong predictions with wrong coarse-grained types.}
\label{fig:distribution}
\end{figure*}

Table~\ref{tab:zeroshot} shows the results on zero-shot entity typing task on \textsc{Few-NERD} dataset. We did not report the performance of the vanilla fine-tuning approach because it cannot produce reasonable results with a randomly initialized classifier. And it also should be noted that the prompt method without fine-tuning already outperforms random guessing.
It indicates that adding a prompt is informative for a model pre-trained on masked-language-model task (e.g. BERT) and can induce reasonable predictions in entity typing tasks. Second, the performance of the model improves by a large margin if trained on unlabeled data. It shows the effectiveness of the proposed self-supervised training approach and points to the potential of a pre-trained prompt-based model under the zero-shot setting when no labeled data are available.

To explore the more subtle changes in performance, we carry out case study for the zero-shot entity typing.
In Figure~\ref{fig:distribution}, we illustrate the zero-shot prediction distribution (the correct prediction and other top-5 predictions) for four entity types in \textsc{Few-NERD}, which are \textsc{Org-SportsTeam},  \textsc{Event-Attack}, \textsc{Misc-Currency} and \textsc{Loc-Mountain}. We could observe that with self-supervised prompt-learning, \textsc{Plet (S)} could summarize entity type information and infer the related words to a certain extent. In Figure~\ref{fig:distribution} (a) and Figure~\ref{fig:distribution} (b), the \textsc{Plet} model suffers from a severe bias and almost predict no correct labels in the zero-shot setting since such words are low-frequency. And although there is no explicit supervision in the pre-training stage of $\textsc{UnPlet}$, the model could still find the corresponding words that express the \textsc{Org-SportsLeague} and the \textsc{Event-Attack} types. In Figure~\ref{fig:distribution} (c), self-supervised learning increases the performance of the original encoder. Further, in Figure~\ref{fig:distribution} (d), \textsc{Plet} has been able to make satisfying predictions for this type \textsc{Loc-Mountain}. In this case, the use of self-supervised learning has hardly weakened the performance, which means that the process of automatically summarizing type information has a little negative impact on high-confidence entity types.


\begin{table*}[!ht]
\centering
\scalebox{0.93}{
\begin{tabular}{cccccc}
\toprule

\textbf{Encoding Strategy} & \multicolumn{2}{l}{\textbf{Template T(x)}} & \textbf{Acc}   & \textbf{MiF}   & \textbf{MaF}   \\ \midrule

\multirow{3}{*}{Hard-encoding}  & \multicolumn{2}{l}{ x.\ \pattern{[\texttt{Ent}]\text{ is }[\texttt{MASK}]}}   &   54.45    &   67.34    &   67.34    \\
 & \multicolumn{2}{l}{ x.\ \pattern{[\texttt{Ent}]\text{ is a }[\texttt{MASK}]}}     &   53.93    &    66.44   &   66.44    \\
& \multicolumn{2}{l}{x.\ \pattern{\text{In this sentence, }[\texttt{E}]\text{ is }[\texttt{MASK}]}}     &   \textbf{55.75}    &   \textbf{68.74}    &     \textbf{68.74}  \\ \midrule
\multirow{4}{*}{Soft-encoding}  & \multicolumn{2}{l}{  x\  \pattern{[\texttt{P}]\  [\texttt{Ent}]\  [\texttt{P}$_1$] ,..., [\texttt{P}$_l$]\  [\texttt{MASK}],} l = 2}     &    \textbf{59.25}   &   \textbf{69.58}    &    \textbf{69.58}   \\
& \multicolumn{2}{l}{  x\  \pattern{[\texttt{P}]\  [\texttt{Ent}]\  [\texttt{P}$_1$],..., [\texttt{P}$_l$]\  [\texttt{MASK}],} l = 3}     &    53.66   &   66.06    &    66.06   \\
& \multicolumn{2}{l}{   x\  \pattern{[\texttt{P}]\  [\texttt{Ent}]\  [\texttt{P}$_1$],..., [\texttt{P}$_l$]\  [\texttt{MASK}],} l = 4}     &   52.96    &   66.01    &   66.01    \\
& \multicolumn{2}{l}{   x\  \pattern{[\texttt{P}]\  [\texttt{Ent}]\  [\texttt{P}$_1$],..., [\texttt{P}$_l$]\  [\texttt{MASK}],} l = 5}     &    55.44   &   68.39    &   68.39   \\ \bottomrule
\end{tabular}}
\caption{Effect of templates. The results are produced under 8-shot setting on \textsc{Few-NERD} dataset by \textsc{Plet}. }
\end{table*}

\subsection{Effect of Templates}
As stated in previous studies~\cite{gao2020making,zhao2021calibrate}, the choice of templates may have a huge impact on the performance in prompt-learning. In this section, we carry out experiments to investigate such influence. Experiments are conducted under the 8-shot setting on \textsc{Few-NERD} dataset, and we use 3 different hard encoding templates and 4 soft encoding templates (by changing the number of prompt tokens $m$).  The results demonstrate that the choice of templates exerts a considerable influence on the performance of prompt-based few-shot learning. For the hard-encoding templates, the phrase that describes the location ``in this sentence'' contributes a remarkable improvement in performance. For the soft-encoding templates, surprisingly, the prompt-learning model yields the best result with the fewest special tokens.


\section{Related Work}

After a series of effective PLMs like GPT~\cite{radfordimproving}, BERT~\cite{devlin2019bert}, RoBERTa~\cite{liu2019roberta} and T5~\cite{raffel2020exploring}, fine-tuned PLMs have demonstrated their effectiveness on various important NLP tasks, such as dialogue generation~\cite{zhang2019dialogpt}, text summarization~\cite{zhang2019pegasus,liu-lapata-2019-text}, question answering~\cite{adiwardana2020humanlike}, and text classification~\cite{baldini-soares-etal-2019-matching,peng2020learning, ding2021prototypical}. 

Despite the success of fine-tuning PLMs, the huge objective form gap between pre-training and fine-tuning still hinders the full use of per-trained knowledge for downstream tasks~\cite{liu2021gpt, han2021ptr, hu2021knowledgeable}. To this end, prompt-learning has been proposed. In prompt-learning, by leveraging language prompts as contexts, downstream tasks can be expressed as some \textit{cloze}-style objectives similar to those pre-training objectives. The seminal work that stimulates the development of prompt-learning is the birth of GPT-3~\cite{brown2020language}, which uses hand-crafted prompts for tuning and achieves very impressive performance on various tasks, especially under the setting of few-shot learning.

Inspired by GPT-3, a series of hand-crafted prompts have been widely explored in knowledge probing~\cite{trinh2018simple,petroni2019language,davison2019commonsense}, relation classification~\cite{han2021ptr}, entiment classification and natural language inference~\cite{schick2020exploiting,liu2021gpt}. To avoid labor-intensive prompt design, automatic prompt search has also been extensively explored~\citet{schick2020automatically,schick2020exploiting,shin2020eliciting,gao2020making, liu2021pre} to generate language phrases for prompts. Recently, some continuous prompts have also been proposed~\cite{li2021prefix,lester2021power}, which directly use a series of learnable continuous embeddings as prompts rather than discrete language phrases.

In this paper, we aim to stimulate PLMs with prompt-learning to capture the attribute information of entities. We take fine-grained entity typing, a crucial task in knowledge extraction to assign entity types to entity mentions~\cite{ling2012fine}, as the foothold to develop prompt-learning strategies. In fact, \citet{dai2021ultra} use hypernym extraction patterns to enhance the context and apply masked language modeling to tackle the ultra-fine entity typing problem~\cite{choi2018ultra} with free-form labels, which shares a similar idea with prompt-learning. In our work, we mainly emphasize using prompt-learning to extract entity types that have been pre-defined in low-data scenarios.



\section{Conclusion}

This work investigates the application of prompt-learning on fine-grained entity typing. More specifically, we proposes a framework $\textsc{Plet}$ that could deal with fine-grained entity typing in fully supervised, few-shot and zero-shot scenarios. 
In \textsc{Plet}, we first introduce a simple and effective prompt-learning pipeline that could be used to extract entity types with both sufficient and insufficient supervision. Furthermore, to handle the zero-shot setting, we propose a self-supervised prompt-learning approach that automatically learns and summarizes entity types based on unlabeled corpora and a pre-defined label schema.  
\textsc{Plet} utilizes prompts to take advantage of prior knowledge distributed in PLMs, and could learn pre-defined type information without overfitting by performing distribution-level optimization. In our future work, along the direction of $\textsc{Plet (S)}$, we will explore better prompt-learning approaches to automatically learning entity types from unlabeled data. 


\bibliographystyle{acl_natbib}
\bibliography{arr,anthology}

\begin{thebibliography}{44}
\expandafter\ifx\csname natexlab\endcsname\relax\def\natexlab#1{#1}\fi

\bibitem[{Adiwardana et~al.(2020)Adiwardana, Luong, So, Hall, Fiedel,
  Thoppilan, Yang, Kulshreshtha, Nemade, Lu et~al.}]{adiwardana2020humanlike}
Daniel Adiwardana, Minh-Thang Luong, David~R So, Jamie Hall, Noah Fiedel, Romal
  Thoppilan, Zi~Yang, Apoorv Kulshreshtha, Gaurav Nemade, Yifeng Lu, et~al.
  2020.
\newblock \href {https://arxiv.org/abs/2001.09977} {Towards a human-like
  open-domain chatbot}.
\newblock \emph{arXiv preprint arXiv:2001.09977}.

\bibitem[{Baldini~Soares et~al.(2019)Baldini~Soares, FitzGerald, Ling, and
  Kwiatkowski}]{baldini-soares-etal-2019-matching}
Livio Baldini~Soares, Nicholas FitzGerald, Jeffrey Ling, and Tom Kwiatkowski.
  2019.
\newblock \href {https://doi.org/10.18653/v1/P19-1279} {Matching the blanks:
  Distributional similarity for relation learning}.
\newblock In \emph{Proceedings of the 57th Annual Meeting of the Association
  for Computational Linguistics}, pages 2895--2905, Florence, Italy.
  Association for Computational Linguistics.

\bibitem[{Brown et~al.(2020)Brown, Mann, Ryder, Subbiah, Kaplan, Dhariwal,
  Neelakantan, Shyam, Sastry, Askell et~al.}]{brown2020language}
Tom~B Brown, Benjamin Mann, Nick Ryder, Melanie Subbiah, Jared Kaplan, Prafulla
  Dhariwal, Arvind Neelakantan, Pranav Shyam, Girish Sastry, Amanda Askell,
  et~al. 2020.
\newblock \href
  {https://proceedings.neurips.cc/paper/2020/file/1457c0d6bfcb4967418bfb8ac142f64a-Paper.pdf}
  {Language models are few-shot learners}.
\newblock In \emph{Proceedings of NIPS}, pages 1877--1901.

\bibitem[{Choi et~al.(2018)Choi, Levy, Choi, and Zettlemoyer}]{choi2018ultra}
Eunsol Choi, Omer Levy, Yejin Choi, and Luke Zettlemoyer. 2018.
\newblock \href {https://aclanthology.org/P18-1009} {Ultra-fine entity typing}.
\newblock In \emph{Proceedings of ACL}, pages 87--96.

\bibitem[{Dai et~al.(2021)Dai, Song, and Wang}]{dai2021ultra}
Hongliang Dai, Yangqiu Song, and Haixun Wang. 2021.
\newblock \href {https://doi.org/10.18653/v1/2021.acl-long.141} {Ultra-fine
  entity typing with weak supervision from a masked language model}.
\newblock In \emph{Proceedings of ACL}, pages 1790--1799.

\bibitem[{Davison et~al.(2019)Davison, Feldman, and
  Rush}]{davison2019commonsense}
Joe Davison, Joshua Feldman, and Alexander~M Rush. 2019.
\newblock \href {https://arxiv.org/abs/1909.00505} {Commonsense knowledge
  mining from pretrained models}.
\newblock In \emph{Proceedings of EMNLP-IJCNLP}, pages 1173--1178.

\bibitem[{Devlin et~al.(2019)Devlin, Chang, Lee, and
  Toutanova}]{devlin2019bert}
Jacob Devlin, Ming-Wei Chang, Kenton Lee, and Kristina Toutanova. 2019.
\newblock \href {https://www.aclweb.org/anthology/N19-1423.pdf} {Bert:
  Pre-training of deep bidirectional transformers for language understanding}.
\newblock In \emph{Proceedings of NAACL-HLT}, pages 4171--4186.

\bibitem[{Ding et~al.(2021{\natexlab{a}})Ding, Wang, Fu, Xu, Wang, Xie, Shen,
  Huang, Zheng, and Zhang}]{ding2021prototypical}
Ning Ding, Xiaobin Wang, Yao Fu, Guangwei Xu, Rui Wang, Pengjun Xie, Ying Shen,
  Fei Huang, Hai-Tao Zheng, and Rui Zhang. 2021{\natexlab{a}}.
\newblock \href {https://openreview.net/forum?id=aCgLmfhIy_f} {Prototypical
  representation learning for relation extraction}.
\newblock In \emph{Proceedings of ICLR}.

\bibitem[{Ding et~al.(2021{\natexlab{b}})Ding, Xu, Chen, Wang, Han, Xie, Zheng,
  and Liu}]{ding2021few}
Ning Ding, Guangwei Xu, Yulin Chen, Xiaobin Wang, Xu~Han, Pengjun Xie, Hai-Tao
  Zheng, and Zhiyuan Liu. 2021{\natexlab{b}}.
\newblock \href {https://aclanthology.org/2021.acl-long.248/} {Few-nerd: A
  few-shot named entity recognition dataset}.
\newblock In \emph{Proceedings of ACL}, pages 3198--3213.

\bibitem[{Gao et~al.(2020)Gao, Fisch, and Chen}]{gao2020making}
Tianyu Gao, Adam Fisch, and Danqi Chen. 2020.
\newblock \href {https://arxiv.org/pdf/2012.15723} {Making pre-trained language
  models better few-shot learners}.
\newblock \emph{arXiv preprint arXiv:2012.15723}.

\bibitem[{Han et~al.(2021{\natexlab{a}})Han, Zhang, Ding, Gu, Liu, Huo, Qiu,
  Zhang, Han, Huang et~al.}]{han2021pre}
Xu~Han, Zhengyan Zhang, Ning Ding, Yuxian Gu, Xiao Liu, Yuqi Huo, Jiezhong Qiu,
  Liang Zhang, Wentao Han, Minlie Huang, et~al. 2021{\natexlab{a}}.
\newblock \href {https://arxiv.org/abs/2106.07139} {Pre-trained models: Past,
  present and future}.
\newblock \emph{arXiv preprint arXiv:2106.07139}.

\bibitem[{Han et~al.(2021{\natexlab{b}})Han, Zhao, Ding, Liu, and
  Sun}]{han2021ptr}
Xu~Han, Weilin Zhao, Ning Ding, Zhiyuan Liu, and Maosong Sun.
  2021{\natexlab{b}}.
\newblock \href {https://arxiv.org/abs/2105.11259} {Ptr: Prompt tuning with
  rules for text classification}.
\newblock \emph{arXiv preprint arXiv:2105.11259}.

\bibitem[{Hewitt and Manning(2019)}]{hewitt2019structural}
John Hewitt and Christopher~D Manning. 2019.
\newblock \href {https://www.aclweb.org/anthology/N19-1419"} {A structural
  probe for finding syntax in word representations}.
\newblock In \emph{Proceedings of NAACL}, pages 4129--4138.

\bibitem[{Hu et~al.(2021)Hu, Ding, Wang, Liu, Li, and
  Sun}]{hu2021knowledgeable}
Shengding Hu, Ning Ding, Huadong Wang, Zhiyuan Liu, Juanzi Li, and Maosong Sun.
  2021.
\newblock \href {https://arxiv.org/abs/2108.02035} {Knowledgeable
  prompt-tuning: Incorporating knowledge into prompt verbalizer for text
  classification}.
\newblock \emph{arXiv preprint arXiv:2108.02035}.

\bibitem[{Jawahar et~al.(2019)Jawahar, Sagot, and Seddah}]{jawahar2019does}
Ganesh Jawahar, Beno{\^\i}t Sagot, and Djam{\'e} Seddah. 2019.
\newblock \href {https://www.aclweb.org/anthology/P19-1356} {What does bert
  learn about the structure of language?}
\newblock In \emph{Proceedings of ACL}, pages 3651--3657.

\bibitem[{Lester et~al.(2021)Lester, Al-Rfou, and Constant}]{lester2021power}
Brian Lester, Rami Al-Rfou, and Noah Constant. 2021.
\newblock \href {https://arxiv.org/abs/2104.08691} {The power of scale for
  parameter-efficient prompt tuning}.
\newblock \emph{arXiv preprint arXiv:2104.08691}.

\bibitem[{Li and Liang(2021)}]{li2021prefix}
Xiang~Lisa Li and Percy Liang. 2021.
\newblock \href {https://arxiv.org/abs/2101.00190} {Prefix-tuning: Optimizing
  continuous prompts for generation}.
\newblock \emph{arXiv preprint arXiv:2101.00190}.

\bibitem[{Lin et~al.(2012)Lin, Etzioni et~al.}]{ling2012fine}
Thomas Lin, Oren Etzioni, et~al. 2012.
\newblock \href {https://dl.acm.org/doi/10.5555/2390948.2391045} {No noun
  phrase left behind: detecting and typing unlinkable entities}.
\newblock In \emph{Proceedings of EMNLP-CoNLL}, pages 893--903.

\bibitem[{Ling and Weld(2012)}]{Ling2012FineGrainedER}
Xiao Ling and Daniel~S. Weld. 2012.
\newblock Fine-grained entity recognition.
\newblock In \emph{AAAI}.

\bibitem[{Liu et~al.(2021{\natexlab{a}})Liu, Yuan, Fu, Jiang, Hayashi, and
  Neubig}]{liu2021pre}
Pengfei Liu, Weizhe Yuan, Jinlan Fu, Zhengbao Jiang, Hiroaki Hayashi, and
  Graham Neubig. 2021{\natexlab{a}}.
\newblock \href {https://arxiv.org/abs/2107.13586} {Pre-train, prompt, and
  predict: A systematic survey of prompting methods in natural language
  processing}.
\newblock \emph{arXiv preprint arXiv:2107.13586}.

\bibitem[{Liu et~al.(2021{\natexlab{b}})Liu, Zheng, Du, Ding, Qian, Yang, and
  Tang}]{liu2021gpt}
Xiao Liu, Yanan Zheng, Zhengxiao Du, Ming Ding, Yujie Qian, Zhilin Yang, and
  Jie Tang. 2021{\natexlab{b}}.
\newblock \href {https://arxiv.org/abs/2103.10385} {Gpt understands, too}.
\newblock \emph{arXiv preprint arXiv:2103.10385}.

\bibitem[{Liu and Lapata(2019)}]{liu-lapata-2019-text}
Yang Liu and Mirella Lapata. 2019.
\newblock \href {https://doi.org/10.18653/v1/D19-1387} {Text summarization with
  pretrained encoders}.
\newblock In \emph{Proceedings of the 2019 Conference on Empirical Methods in
  Natural Language Processing and the 9th International Joint Conference on
  Natural Language Processing (EMNLP-IJCNLP)}, pages 3730--3740, Hong Kong,
  China. Association for Computational Linguistics.

\bibitem[{Liu et~al.(2019)Liu, Ott, Goyal, Du, Joshi, Chen, Levy, Lewis,
  Zettlemoyer, and Stoyanov}]{liu2019roberta}
Yinhan Liu, Myle Ott, Naman Goyal, Jingfei Du, Mandar Joshi, Danqi Chen, Omer
  Levy, Mike Lewis, Luke Zettlemoyer, and Veselin Stoyanov. 2019.
\newblock \href {https://arxiv.org/abs/1907.11692} {Roberta: A robustly
  optimized bert pretraining approach}.
\newblock \emph{arXiv preprint arXiv:1907.11692}.

\bibitem[{Loshchilov and Hutter(2019)}]{loshchilov2018decoupled}
Ilya Loshchilov and Frank Hutter. 2019.
\newblock \href {https://openreview.net/forum?id=Bkg6RiCqY7} {Decoupled weight
  decay regularization}.
\newblock In \emph{Proceedings of ICLR}.

\bibitem[{Paszke et~al.(2019)Paszke, Gross, Massa, Lerer, Bradbury, Chanan,
  Killeen, Lin, Gimelshein, Antiga, Desmaison, K{\"{o}}pf, Yang, DeVito,
  Raison, Tejani, Chilamkurthy, Steiner, Fang, Bai, and
  Chintala}]{paszke2019pytorch}
Adam Paszke, Sam Gross, Francisco Massa, Adam Lerer, James Bradbury, Gregory
  Chanan, Trevor Killeen, Zeming Lin, Natalia Gimelshein, Luca Antiga, Alban
  Desmaison, Andreas K{\"{o}}pf, Edward Yang, Zachary DeVito, Martin Raison,
  Alykhan Tejani, Sasank Chilamkurthy, Benoit Steiner, Lu~Fang, Junjie Bai, and
  Soumith Chintala. 2019.
\newblock \href
  {https://proceedings.neurips.cc/paper/2019/hash/bdbca288fee7f92f2bfa9f7012727740-Abstract.html}
  {Pytorch: An imperative style, high-performance deep learning library}.
\newblock In \emph{Proceedings of NIPS}, pages 8024--8035.

\bibitem[{Peng et~al.(2020)Peng, Gao, Han, Lin, Li, Liu, Sun, and
  Zhou}]{peng2020learning}
Hao Peng, Tianyu Gao, Xu~Han, Yankai Lin, Peng Li, Zhiyuan Liu, Maosong Sun,
  and Jie Zhou. 2020.
\newblock \href {https://doi.org/10.18653/v1/2020.emnlp-main.298} {{L}earning
  from {C}ontext or {N}ames? {A}n {E}mpirical {S}tudy on {N}eural {R}elation
  {E}xtraction}.
\newblock In \emph{Proceedings of EMNLP}, pages 3661--3672.

\bibitem[{Petroni et~al.(2019)Petroni, Rockt{\"a}schel, Riedel, Lewis, Bakhtin,
  Wu, and Miller}]{petroni2019language}
Fabio Petroni, Tim Rockt{\"a}schel, Sebastian Riedel, Patrick Lewis, Anton
  Bakhtin, Yuxiang Wu, and Alexander Miller. 2019.
\newblock \href {https://www.aclweb.org/anthology/D19-1250.pdf} {Language
  models as knowledge bases?}
\newblock In \emph{Proceedings of EMNLP}, pages 2463--2473.

\bibitem[{Qiu et~al.(2020)Qiu, Sun, Xu, Shao, Dai, and Huang}]{qiu2020pre}
Xipeng Qiu, Tianxiang Sun, Yige Xu, Yunfan Shao, Ning Dai, and Xuanjing Huang.
  2020.
\newblock \href {https://arxiv.org/abs/2003.08271v2} {Pre-trained models for
  natural language processing: A survey}.
\newblock \emph{Science China Technological Sciences}, pages 1--26.

\bibitem[{Radford et~al.(2018)Radford, Narasimhan, Salimans, and
  Sutskever}]{radfordimproving}
Alec Radford, Karthik Narasimhan, Tim Salimans, and Ilya Sutskever. 2018.
\newblock \href
  {https://www.cs.ubc.ca/~amuham01/LING530/papers/radford2018improving.pdf}
  {Improving language understanding by generative pre-training}.
\newblock \emph{OpenAI}.

\bibitem[{Raffel et~al.(2020)Raffel, Shazeer, Roberts, Lee, Narang, Matena,
  Zhou, Li, and Liu}]{raffel2020exploring}
Colin Raffel, Noam Shazeer, Adam Roberts, Katherine Lee, Sharan Narang, Michael
  Matena, Yanqi Zhou, Wei Li, and Peter~J Liu. 2020.
\newblock \href {https://www.jmlr.org/papers/volume21/20-074/20-074.pdf}
  {Exploring the limits of transfer learning with a unified text-to-text
  transformer}.
\newblock \emph{JMLR}, 21:1--67.

\bibitem[{Ren et~al.(2016{\natexlab{a}})Ren, He, Qu, Huang, Ji, and
  Han}]{ren-etal-2016-afet}
Xiang Ren, Wenqi He, Meng Qu, Lifu Huang, Heng Ji, and Jiawei Han.
  2016{\natexlab{a}}.
\newblock \href {https://doi.org/10.18653/v1/D16-1144} {{AFET}: Automatic
  fine-grained entity typing by hierarchical partial-label embedding}.
\newblock In \emph{Proceedings of the 2016 Conference on Empirical Methods in
  Natural Language Processing}, pages 1369--1378, Austin, Texas. Association
  for Computational Linguistics.

\bibitem[{Ren et~al.(2016{\natexlab{b}})Ren, He, Qu, Voss, Ji, and
  Han}]{ren2016label}
Xiang Ren, Wenqi He, Meng Qu, Clare~R. Voss, Heng Ji, and Jiawei Han.
  2016{\natexlab{b}}.
\newblock \href {https://doi.org/10.1145/2939672.2939822} {Label noise
  reduction in entity typing by heterogeneous partial-label embedding}.
\newblock In \emph{Proceedings of SIGKDD}, page 1825–1834.

\bibitem[{Schick et~al.(2020)Schick, Schmid, and
  Sch{\"u}tze}]{schick2020automatically}
Timo Schick, Helmut Schmid, and Hinrich Sch{\"u}tze. 2020.
\newblock \href {https://doi.org/10.18653/v1/2020.coling-main.488}
  {Automatically identifying words that can serve as labels for few-shot text
  classification}.
\newblock In \emph{Proceedings of COLING}, pages 5569--5578.

\bibitem[{Schick and Sch{\"u}tze(2021)}]{schick2020exploiting}
Timo Schick and Hinrich Sch{\"u}tze. 2021.
\newblock \href {https://www.aclweb.org/anthology/2021.eacl-main.20}
  {Exploiting cloze-questions for few-shot text classification and natural
  language inference}.
\newblock In \emph{Proceedings of EACL}, pages 255--269.

\bibitem[{Shimaoka et~al.(2017)Shimaoka, Stenetorp, Inui, and
  Riedel}]{shimaoka-etal-2017-neural}
Sonse Shimaoka, Pontus Stenetorp, Kentaro Inui, and Sebastian Riedel. 2017.
\newblock \href {https://www.aclweb.org/anthology/E17-1119} {Neural
  architectures for fine-grained entity type classification}.
\newblock In \emph{Proceedings of the 15th Conference of the {E}uropean Chapter
  of the Association for Computational Linguistics: Volume 1, Long Papers},
  pages 1271--1280, Valencia, Spain. Association for Computational Linguistics.

\bibitem[{Shin et~al.(2020)Shin, Razeghi, Logan~IV, Wallace, and
  Singh}]{shin2020eliciting}
Taylor Shin, Yasaman Razeghi, Robert~L Logan~IV, Eric Wallace, and Sameer
  Singh. 2020.
\newblock \href {https://www.aclweb.org/anthology/2020.emnlp-main.346.pdf}
  {Autoprompt: Eliciting knowledge from language models using automatically
  generated prompts}.
\newblock In \emph{Proceedings of EMNLP}, pages 4222--4235.

\bibitem[{Trinh and Le(2018)}]{trinh2018simple}
Trieu~H Trinh and Quoc~V Le. 2018.
\newblock \href {https://arxiv.org/abs/1806.02847} {A simple method for
  commonsense reasoning}.
\newblock \emph{arXiv preprint arXiv:1806.02847}.

\bibitem[{Wang et~al.(2021)Wang, Ding, Li, and Zheng}]{wang2021cline}
Dong Wang, Ning Ding, Piji Li, and Haitao Zheng. 2021.
\newblock \href {https://aclanthology.org/2021.acl-long.181} {{CLINE}:
  Contrastive learning with semantic negative examples for natural language
  understanding}.
\newblock In \emph{Proceedings of ACL}.

\bibitem[{Weischedel and Brunstein(2005)}]{Weischedel2005}
Ralph Weischedel and Ada Brunstein. 2005.
\newblock \href {https://doi.org/https://doi.org/10.35111/9fx9-gz10} {{BBN
  Pronoun Coreference and Entity Type Corpus}}.
\newblock Linguistic Data Consortium, Philadelphia.

\bibitem[{Weischedel et~al.(2013)Weischedel, Palmer, Marcus, Hovy, Pradhan,
  Ramshaw, Xue, Taylor, Kaufman, Franchini, El-Bachouti, Belvin, and
  Houston}]{MKJJ2R_2013}
Ralph Weischedel, Martha Palmer, Mitchell Marcus, Eduard Hovy, Sameer Pradhan,
  Lance Ramshaw, Nianwen Xue, Ann Taylor, Jeff Kaufman, Michelle Franchini,
  Mohammed El-Bachouti, Robert Belvin, and Ann Houston. 2013.
\newblock \href {https://doi.org/11272.1/AB2/MKJJ2R} {{OntoNotes Release 5.0}}.
\newblock Abacus Data Network.

\bibitem[{Wolf et~al.(2020)Wolf, Debut, Sanh, Chaumond, Delangue, Moi, Cistac,
  Rault, Louf, Funtowicz, Davison, Shleifer, von Platen, Ma, Jernite, Plu, Xu,
  Le~Scao, Gugger, Drame, Lhoest, and Rush}]{wolf2019huggingface}
Thomas Wolf, Lysandre Debut, Victor Sanh, Julien Chaumond, Clement Delangue,
  Anthony Moi, Pierric Cistac, Tim Rault, Remi Louf, Morgan Funtowicz, Joe
  Davison, Sam Shleifer, Patrick von Platen, Clara Ma, Yacine Jernite, Julien
  Plu, Canwen Xu, Teven Le~Scao, Sylvain Gugger, Mariama Drame, Quentin Lhoest,
  and Alexander Rush. 2020.
\newblock \href {https://www.aclweb.org/anthology/2020.emnlp-demos.6}
  {Transformers: State-of-the-art natural language processing}.
\newblock In \emph{Proceedings of EMNLP}, pages 38--45.

\bibitem[{Zhang et~al.(2019)Zhang, Zhao, Saleh, and Liu}]{zhang2019pegasus}
Jingqing Zhang, Yao Zhao, Mohammad Saleh, and Peter~J. Liu. 2019.
\newblock \href {https://arxiv.org/abs/1912.08777} {Pegasus: Pre-training with
  extracted gap-sentences for abstractive summarization}.
\newblock In \emph{Proceedings of ICML}, pages 11328--11339.

\bibitem[{Zhang et~al.(2020)Zhang, Sun, Galley, Chen, Brockett, Gao, Gao, Liu,
  and Dolan}]{zhang2019dialogpt}
Yizhe Zhang, Siqi Sun, Michel Galley, Yen-Chun Chen, Chris Brockett, Xiang Gao,
  Jianfeng Gao, Jingjing Liu, and Bill Dolan. 2020.
\newblock \href {https://arxiv.org/abs/1911.00536} {Dialogpt: Large-scale
  generative pre-training for conversational response generation}.
\newblock In \emph{Proceedings of ACL}, pages 270--278.

\bibitem[{Zhao et~al.(2021)Zhao, Wallace, Feng, Klein, and
  Singh}]{zhao2021calibrate}
Tony~Z Zhao, Eric Wallace, Shi Feng, Dan Klein, and Sameer Singh. 2021.
\newblock \href {https://arxiv.org/pdf/2102.09690.pdf} {Calibrate before use:
  Improving few-shot performance of language models}.
\newblock \emph{arXiv preprint arXiv:2102.09690}.

\end{thebibliography}

\end{document}